\documentclass[11pt]{article}

\usepackage[final]{acl}

\usepackage{times}
\usepackage{latexsym}

\usepackage[T1]{fontenc}

\usepackage[utf8]{inputenc}

\usepackage{microtype}

\usepackage{inconsolata}

\usepackage{graphicx}

\usepackage{amsmath}
\usepackage{amssymb}
\usepackage{xcolor}
\usepackage{multirow}
\usepackage{tikz}
\usetikzlibrary{calc}
\usepackage[capitalise]{cleveref}
\usepackage[all]{foreign}

\newtheorem{definition}{Definition}
\DeclareMathOperator{\V}{\mathcal{V}}
\DeclareMathOperator{\C}{\mathcal{C}}
\DeclareMathOperator{\R}{\mathcal{R}}
\DeclareMathOperator{\F}{\mathcal{F}}

\DeclareMathOperator{\D}{\mathcal{D}}
\DeclareMathOperator{\Cube}{Cube}
\DeclareMathOperator{\Tet}{Tet}
\DeclareMathOperator{\LeftOf}{LeftOf}

\usepackage{xcolor}
\usepackage{booktabs}
\usepackage{tabularx}
\usepackage{graphicx}

\title{Natural Language to First-Order Logic LLM-based Autoformalization}

\author{
  Andrea Brunello$^1$ \quad
  Cristian Curaba$^1$ \quad
  Luca Geatti$^1$ \AND
  Michele Mignani$^{1*}$ \quad
  Angelo Montanari$^1$ \quad
  Nicola Saccomanno$^1$ \\[0.5em]
  $^1$University of Udine, Italy \\
  \texttt{name.surname@uniud.it} \\[0.3em]
  {\small $^*$Corresponding author.}
}

\begin{document}
\maketitle
\begin{abstract}
Large Language Models (LLMs) have renewed interest in autoformalization. Yet, when First-Order Logic (FOL) is considered as the target formalism, the field still lacks a unified task formulation and a systematic survey. This paper addresses this gap: we first provide a principled definition for the FOL-autoformalization task by distinguishing Ontology Extraction from Logical Translation, showing how their conflation obscures (cross-study) evaluation; we review existing datasets, evaluation metrics, and LLM-based methods, including fine-tuning, prompting, and verification-based refinement; we identify open challenges in benchmarking, semantic evaluation, ontology-aware methods, and end-to-end applications.  
\end{abstract}  

\section{Introduction}

In the literature, autoformalization is broadly defined as the task of automatically generating statements in a target formalism that capture the semantics of a Natural Language (NL) sentence~\cite{DBLP:conf/mkm/Szegedy20,DBLP:conf/aaai/MensfeltCFKTS26,DBLP:conf/nips/WuJLRSJS22}. Although NL is well-suited for everyday human communication, its flexibility, expressiveness, and inherent ambiguity make it ill-suited for interaction with automated reasoning tools, which require inputs to be unambiguous and specified within a formal semantic framework. Early work relied on rule-based systems~\cite{kamath2018survey} and controlled natural languages~\cite{kuhn2014survey}, while more recent approaches have adopted neural network-based methods~\cite{DBLP:journals/symmetry/JiangC24}. 
Despite the considerable research effort devoted to autoformalization, the field still lacks a unified formulation and a systematic survey when First-Order Logic (FOL) is considered as the target formalism. Here, we address this gap.

\paragraph{Why this survey.}
The motivations are threefold. 
First, 
Large Language Models (LLMs) have reshaped the autoformalization research landscape: their flexibility and general-purpose linguistic competence make autoformalization feasible across a wide range of domains~\cite{DBLP:conf/aaai/MensfeltCFKTS26}.
Second, no systematic survey of LLM-based autoformalization for FOL currently exists, despite FOL offering a well-established balance between expressive power and algorithmic tractability. Surveys for other formalisms, such as Linear Temporal Logic~\cite{DBLP:journals/access/GerminianiNP25} and SQL~\cite{DBLP:journals/corr/abs-2410-01066}, already play a pivotal role in their respective landscapes. 
Third, autoformalization underpins a wide range of relevant applications. In reasoning tasks, symbolic methods such as theorem provers and Satisfiability Modulo Theories (SMT) solvers are reliable and explainable, but require inputs in a precise logical format; autoformalization is thus crucial for converting NL task instances into appropriate 
representations. Recent work shows that inserting a formalization step and invoking symbolic tools can outperform directly prompting an LLM to solve multi-step logical reasoning problems~\cite{DBLP:conf/nips/YeCDD23,DBLP:conf/emnlp/OlaussonGLZSTL23,DBLP:conf/emnlp/PanAWW23,DBLP:journals/corr/abs-2501-14540}, with FOL yielding the best performance in certain settings~\cite{DBLP:conf/i-semantics/BeiserMP25}. Beyond reasoning, reliable formalization tools could make formal verification accessible to non-experts and enable scalable verification of large text corpora, an especially important prospect for AI safety approaches that regard formal guarantees as essential~\cite{Toward_guaranteed_safe_AI}. The significance of autoformalization in FOL has also been demonstrated in logical fallacy detection~\cite{lalwani2024autoformalizing}, argumentation mining~\cite{sun-etal-2025-learning}, legal reasoning~\cite{Alam2023DataAK}, and anomaly detection~\cite{DBLP:conf/aaai/JinFMLDSS25}.

\paragraph{Our contribution.} We provide a detailed description of the FOL-autoformalization task (Section~\ref{sec:task-def}) aimed at resolving existing ambiguities. Surveying the existing literature, we then focus on \textit{evaluation}, covering datasets and metrics (Section~\ref{sec:evaluation}), and \textit{methods} proposed to achieve autoformalization via LLMs (Section~\ref{sec:enhance}). Finally, we highlight \textit{open problems} and directions for future research (Section~\ref{sec:open-gaps}).

\section{Autoformalization Definition}
\label{sec:task-def}

Despite a growing body of work on FOL-autoformalization---both as an engineering component and as a task of independent interest---the field lacks a shared, precise formulation of what the task requires. Informal descriptions typically present autoformalization as the translation of a NL text into a FOL formula that preserves its semantic content. This formulation is 
 ambiguous: it leaves open what formal machinery the formula is interpreted over, and it does not specify under what conditions a formula counts as correct or faithful.

In this section we give a more precise account. We argue that any formalization in FOL is relative to an \emph{ontology}---a structure comprising a formal vocabulary, a labelling of that vocabulary with intended NL meanings, and a background theory. The \emph{raw translation} \(p \in \text{NL} \mapsto \varphi \in \text{FOL}\), typically found in the literature, is therefore an abbreviation for a richer artifact. We then introduce Definition \ref{def-auto} for the FOL-autformalization task-

We provide FOL background in Appendix~\ref{sec:fol_formalism}.

\subsection{Formulas and Ontological Commitments}

Consider the sentence:
$p :=$ ``The cube $A$ is on the left of the tetrahedron $B$.''
A natural FOL candidate is:
$\varphi :=  \mathsf{Cube}(A) \land \mathsf{Tet}(B) \land \mathsf{LeftOf}(A,B)$;
%
\(\varphi\)  contains the non-logical symbols
$\{A,\, B,\, \mathsf{Cube}/1,\, \mathsf{Tet}/1,\, \mathsf{LeftOf}/2\}$,
which jointly constitute the \emph{signature} \(\sigma_\varphi\) that \(\varphi\) is written over. The signature is derivable from the formula alone.

What is not derivable from \(\varphi\) alone is the \emph{intended meaning} of these symbols. Nothing in 
\(\varphi\) determines that \(\mathsf{Cube}(x)\) means ``\(x\) is a cube'' rather than, say, ``\(x\) is a room'' or any other arbitrary unary property. The suggestive symbol names give a strong pragmatic hint, but hints are not meanings: a formula with symbols renamed
$\mathsf{P}_1(c_1) \land \mathsf{P}_2(c_2) \land \mathsf{R}(c_1,c_2)$
is logically identical to \(\varphi\), yet gives no hint at all.

This has a direct consequence for evaluation. Two formulas \(\varphi\) and \(\varphi'\) may disagree not only in logical structure but also in what the symbols they use are intended to mean. 
Without fixing the intended meanings, the problem of comparing formalizations---or judging whether any single one is faithful---is not well-defined.

\subsection{Ontology: Signature, Labelling, Theory}
\label{sec:ontology}

We use the term \emph{ontology} for the structure relative to which a FOL formula represents NL content. We define it as a triple:
$\Omega = \langle \sigma, L, T \rangle$,
whose components are as follows.

\paragraph{Signature \(\sigma\).}
A FOL signature is a set of non-logical symbols with their arities: constants, function symbols, and predicate symbols. Signature \(\sigma\) specifies the vocabulary available for translation. In the running example:
$\sigma = \{A,\, B,\, \mathsf{Cube}/1,\, \mathsf{Tet}/1,\, \mathsf{LeftOf}/2\}$.

\paragraph{Labelling \(L\).}
The labelling is a possibly partial function that maps each symbol \(s \in \sigma\) to an NL meaning-specification \(L(s)\). For our $p$, for instance,  
$L(\mathsf{Cube})$ = \text{``}x\text{ is a cube''},
When \(L\) is fully specified, we call \(\langle \sigma, L \rangle\) a \emph{labelled signature}. It is the labelled signature---not the signature alone---that permits assessment of whether \(\varphi\) faithfully formalizes \(p\). \emph{Raw translation} output exposes \(\sigma_\varphi\) but typically leaves \(L\) implicit, conventional, or underspecified.

\paragraph{Theory \(T\).}
The theory is a possibly empty set of FOL sentences over \(\sigma\) that encode background axioms, taxonomic relations, and domain-specific assumptions. In our example, the formula \lq\lq $ \forall x\,(\mathsf{Cube}(x) \to \neg \mathsf{Tet}(x))$\rq\rq{} that expresses that no object is simultaneously a cube and a tetrahedron, can be an element of $T$.
 In many sentence-level datasets, \(T\) is empty or left implicit. When autoformalization feeds into reasoning or verification, however, the theory is often essential: a formula that is faithful relative to one theory may be inadequate relative to another.
\footnote{The term ``ontology'' is used inconsistently in the literature: some work uses it to mean what we call a labelled signature~\cite{DBLP:conf/aaai/BrunelloGMMS26}, while knowledge representation and ontology learning more commonly use it to include axioms as well~\cite{ARMARY2025100693,du2024short}. We adopt the broader usage, and refer explicitly to the labelled signature or the theory when the distinction matters.}

\subsection{(Not) Uniqueness of Ontological Choices 
}

A single NL sentence does not necessary determine a unique formalization---due to ambiguity, synonymy, or granularity choices---even relative to a fixed ontology. It also does not determine a unique ontology. Consider
$p' :=$ ``Every student passed some exam.''
Three candidate formalizations are:
$\varphi_1 =
\forall x\,\bigl(\mathsf{Student}(x) \to
\exists y\,(\mathsf{Exam}(y) \land \mathsf{Passed}(x,y))\bigr),$ 
$\ \varphi_2 =
\forall x\,\bigl(\mathsf{Student}(x) \to \mathsf{PassedSomeExam}(x)\bigr),$
$\varphi_3 =
\forall x\,\bigl(\mathsf{Student}(x) \to
\exists y\,\exists g\,(\mathsf{Exam}(y) \land \mathsf{Grade}(x,y,g) \land \mathsf{Sufficient}(g))\bigr).$

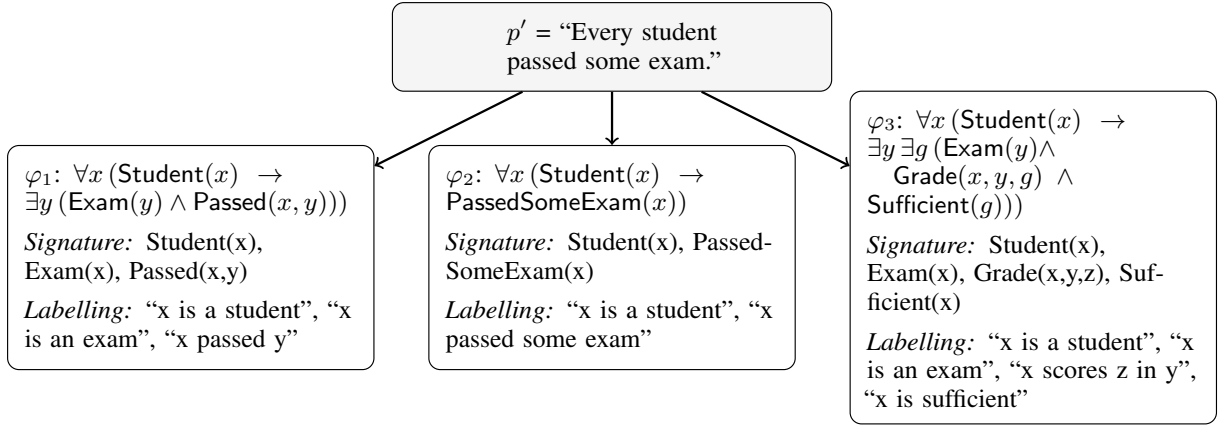
\begin{figure*}[t]
\centering
\resizebox{\textwidth}{!}{%
\begin{tikzpicture}[
  every node/.style={font=\small},
  box/.style={draw, rounded corners, align=left, text width=4.1cm, inner sep=6pt},
  pbox/.style={draw, rounded corners, align=center, text width=5cm, inner sep=6pt, fill=gray!8},
  arr/.style={->, thick}
]

\node[pbox] (p) at (0,0) {$p'$ = ``Every student passed some exam.''};

\node[box] (phi1) at (-5.2,-2.6)
  {$\varphi_1$: $\forall x\,(\mathsf{Student}(x)\to$ $\exists y\,(\mathsf{Exam}(y)\land\mathsf{Passed}(x,y)))$\\[4pt]
   \textit{Signature:} Student(x), Exam(x), Passed(x,y) \\[4pt]
   \textit{Labelling:} \lq\lq x is a student\rq\rq, \lq\lq x is an exam\rq\rq, \lq\lq x passed y\rq\rq};

\node[box] (phi2) at (0,-2.6)
  {$\varphi_2$: $\forall x\,(\mathsf{Student}(x)\to$ $\mathsf{PassedSomeExam}(x))$\\[4pt]
   \textit{Signature:} Student(x), PassedSomeExam(x) \\[4pt]
   \textit{Labelling:} \lq\lq x is a student\rq\rq, \lq\lq x passed some exam\rq\rq};

\node[box] (phi3) at (5.2,-2.6)
  {$\varphi_3$: $\forall x\,(\mathsf{Student}(x)\to$ $\exists y\,\exists g\,(\mathsf{Exam}(y)\land$\\
   \hspace{1em}$\mathsf{Grade}(x,y,g)\land\mathsf{Sufficient}(g)))$\\[4pt]
   \textit{Signature:} Student(x), Exam(x), Grade(x,y,z), Sufficient(x) \\[4pt]
    \textit{Labelling:} \lq\lq x is a student\rq\rq, \lq\lq x is an exam\rq\rq, \lq\lq x scores z in y\rq\rq, \lq\lq x is sufficient\rq\rq};

\draw[arr] (p) -- (phi1);
\draw[arr] (p) -- (phi2);
\draw[arr] (p) -- (phi3);

\end{tikzpicture}%
}
\caption{A single NL sentence $p'$ admits multiple formalizations $\varphi_1,\varphi_2,\varphi_3$, each committing to a different signature. The three formulas can be considered all correct but they express information differently. }
\label{fig:phi_1}
\end{figure*}

From a formal perspective, these are not syntactic variants: each relies on a different signature, labeling, and potentially a different background theory (see Figure \ref{fig:phi_1}). 
Formula $\varphi_1$ represents passing as a binary relation between individuals; $\varphi_2$ absorbs both the exam and the passing event into a single unary predicate; $\varphi_3$ introduces grades as a refined object. Crucially, the choice of signature is not merely a matter of style: it determines both what can be represented about the NL sentence $p'$ and what can subsequently be inferred from its formalization. For instance, $\varphi_2$ cannot adequately represent information about \emph{which} exam a student passed, since the exam is not represented as an individual but is compiled away inside the unary predicate; similarly, neither $\varphi_1$ nor $\varphi_2$ can represent information about a student's grade, since grades are absent from their signatures entirely. These are not failures of translation but direct consequences of the ontological commitments each formula adopts.

In the absence of further information, there is no principled reason to prefer one signature over another; however, these differing ontological commitments make some formalizations more suitable than others depending on the context in which they will be used. For instance, if $p'$ is expected to be only one sentence within a larger corpus concerning different kinds of exams, the ontology underlying $\varphi_2$ is likely inadequate, since it conflates the concept of an exam with the act of a student passing it; a signature in which the exam is reified as an independent object would be preferable. Likewise, if other sentences in the corpus explicitly discuss grades, the ontology underlying $\varphi_3$ is the best-suited of the three considered here, since it is the only one capable of representing and reasoning about that information at all.

This observation implies that evaluating autoformalization systems by comparing formulas directly---without accounting for the ontologies they presuppose---conflates two distinct sources of disagreement: different translations of the same NL content, and different ontological choices about how to represent that content at all.

\subsection{FOL-autoformalization Definition}

We provide now a precise definition of the task.

\begin{definition}[FOL-autoformalization task]
\label{def-auto}
Let $p$ be a NL text, and let $K$ be optional background knowledge given as input, which may
include, e.g., contextual information or a (partial) specification of an ontology
$\Tilde{\Omega} = \langle \Tilde{\sigma}, \Tilde{L}, \Tilde{T} \rangle$.

FOL-autoformalization is the task of automatically producing, from $\langle p, K \rangle$,
a pair $\langle \varphi, \Omega \rangle$, with  $\Omega = \langle \sigma, L, T \rangle$, where $L$ and $T$ may possibly be empty, i.e., left implicit.
 Specifically, $\varphi$ is a FOL formula over $\sigma$ that represents the content of $p$ relative to $\Omega$: under the meanings assigned by $L$ and the constraints imposed by $T$, $\varphi$ holds if and only if $p$ is true.
\end{definition} 

The components of $\Omega$ may be given as output or left implicit.
The signature $\sigma$ is always recoverable
from $\varphi$ itself. The labelling function
$L$, when omitted, leaves the intended NL meaning of the symbols in $\sigma$
implicit: the grounding of the formula to natural
language must then be inferred from context. Similarly, the background theory $T$,
when omitted, imposes no additional constraints on the interpretation of $\sigma$,
potentially leaving the semantics of $\varphi$ underspecified. The richer the
specification of $\Omega$, the more verifiable the output $\varphi$ becomes.

Definition \ref{def-auto} is deliberately general: different task variants arise by
varying what $K$ is and which components of $\Omega$ are
required to be explicit. Two extremes are worth noting. At one end,
\emph{raw translation}---the most common variant in the literature---doesn't consider $\Tilde{\Omega}$ in $K$ and requires only $\varphi$
as output, leaving $L$ and $T$ implicit. At the other end,
\emph{full autoformalization} requires producing both $\varphi$ and the
complete $\Omega$, and is the most theoretically complete instantiation of the definition. The strength of this definition lies precisely in its ability to capture both extremes---and everything in between---within a single unified framework.

Another variant present in the literature (see \citet{DBLP:conf/aaai/BrunelloGMMS26}) is the \emph{Logical Translation} (LT), where the input context $K$ contains an ontology $\Omega = \langle \sigma, L\rangle$, and the output is the formula $\varphi$ (and the same $\Omega$ provided in input). LT is relevant when a domain ontology already exists and the goal is to produce formalizations compatible with it. This frequently occurs in formal verification pipelines or multi-sentence corpora requiring consistent predicate use across formulas.

Furthermore, LT is important since it provides a useful decomposition of the \emph{full autoformalization} task, as noted in \cite{DBLP:conf/aaai/BrunelloGMMS26}. Since LT assumes $\Omega$, they propose to consider the complementary task of \emph{Ontology Extraction} (OE):
 given contextual information and the NL text $p$, the goal is to construct the specific ontology $\Omega$ required to formalize $p$ adequately. OE becomes the critical subtask whenever a reusable ontology must be established before LT can proceed. For instance, it is needed when bootstrapping a domain vocabulary from a set of NL documents (legal texts, engineering specifications) that will later serve as the basis for formalizing arbitrary sentences in that domain.

This decomposition is not the only possible one, but it gives dignity to the distinct nature of these subtasks, the separate sources of error they
introduce, and the different communities whose methods are relevant to each (Ontology Learning and Autoformalization).
As we discuss in Section~\ref{sec-metrics}, it also enables theoretically
grounded evaluation.

\section{Autoformalization Evaluation} \label{sec:evaluation}
Here we explore the datasets and metrics used in the literature for evaluation, deferring their use in performance-enhancement methods to Section~\ref{sec:enhance}.

\paragraph{Ontology Evaluation.}
In principle, evaluation should assess both the quality of the ontology and that of the resulting FOL formula. In practice, no existing work explicitly evaluates signature quality, even when the OE/LT decomposition is acknowledged---a gap shared with ontology learning, where evaluation is also conducted indirectly via downstream tasks~\cite{du2024short}. In FOL-autoformalization, signature quality is therefore assessed only implicitly: a correct formula is taken as evidence that the employed symbols constitute an adequate signature. Similarly for the theory, that is considered by only a small portion of the FOL-autoformalization literature~\cite{lalwani2024autoformalizing,DBLP:journals/corr/abs-2501-14540}, and substantially overlaps with the field of ontology learning and commonsense knowledge acquisition~\cite{ARMARY2025100693,du2024short}, which have their own methods and evaluation frameworks.  
Accordingly, we focus on the evaluation of FOL formulas: this can be done either directly or via proxy tasks.

\paragraph{Evaluation via Proxy Tasks.}
A fundamental challenge is the cost of obtaining ground-truth formulas: correct FOL annotations require significant expertise, making large-scale direct evaluation often prohibitive. Proxy tasks offer a practical alternative, measuring whether the formalization succeeds at a downstream application rather than against a ground truth. Any application can serve this role, provided formalization success is clearly reflected in the overall outcome; for example, \citet{lalwani2024autoformalizing} use formalization for logical fallacy detection, while \citet{karia2024forallutoexistsvalautonomousassessmentllms} formalize a NL sentence back from a FOL formula and checks equivalence with the original.

In practice, though, most of the literature uses \emph{Natural Language Inference} (NLI) task as a proxy. In NLI, given premises $p_1, \dots, p_n$ and a conclusion $c$, the task is to determine whether $c$ is logically entailed by the premises. This entailment relation should be preserved under formalization: if $p_1, \dots, p_n, c$ are formalized as $\varphi_1, \dots, \varphi_n, \psi$, then $c$ is entailed by (resp., contradicts) the premises $p_1, \dots, p_n$ if and only if $\psi$ (resp., $\neg \psi$) is derivable from the premises $\varphi_1, \dots, \varphi_n$. 
This methodology has well-known limitations \cite{DBLP:conf/aaai/BrunelloGMMS26}: entailment agreement between NL and FOL can occur even with incorrect formulas; a mismatch can arise from any subset of incorrectly formalized sentences; and NL entailment may rely on background knowledge not captured by the formalization.
\footnote{Some limitations can be mitigated by generating multiple conclusions depending on different premise subsets, or by using datasets that do not require background knowledge.} 

We focus on direct evaluation here, and refer the reader to \citet{DBLP:conf/naacl/MadaanESPH25} for a dedicated treatment of the NLI setting.

    \subsection{Datasets}
    \label{sec:datasets}
    We review how existing datasets pairing a NL sentence to a FOL formula (NL–FOL) are constructed and validated. We discuss FOLIO~\cite{DBLP:conf/emnlp/HanS0QRZCPQBSWS24}, Willow~\cite{deveci_2024}, MALLS~\cite{DBLP:conf/acl/YangXPSF24}, ProverQA~\cite{DBLP:conf/iclr/QiM0DHWLH25}, PROOFFOL~\cite{DBLP:journals/corr/abs-2409-16461}, Text2log~\cite{9401852}, and the private (but available on request
    ) Grade Grinder Corpus (GGC)~\cite{barker2011student}. A high-level summary is provided in Table~\ref{tab:dataset-simple}. 
    Purely NLI datasets without a validated FOL layer are excluded from the main body of the paper, as they are not directly usable for autoformalization evaluation. Further information about these datasets, and what would need to be supplied to make them usable for the autoformalization task, is provided in Appendix \ref{app:reasoning-datasets}.

\begin{table}[t]
\centering
\resizebox{\columnwidth}{!}{
\begin{tabular}{lcccc}
\hline
\multirow{2}{*}{Dataset}
& \multicolumn{2}{c}{Generation Type} 
& \multirow{2}{*}{No. of Pairs}\\
\cline{2-3}
& NL & FOL &  \\
\hline
FOLIO 
& Human 
& Human (*)
& $\sim$3800\\

GGC 
& Human
& Human (*)

 & $\sim$280\\

MALLS 
& LLM 
& LLM 
& $\sim$28K \\

ProverQA 
& LLM (*)
& Rule-based 
& $\sim$17K \\

Willow 
& LLM 
& LLM 
& $\sim$16K \\

PROOFFOL 
& Human 
& LLM (*) 
& $\sim$10K \\

Text2log 
& Rule-based 
& Rule-based (*)  
& $\sim$100K\\
\hline
\end{tabular}
}
\caption{Overview of NL–FOL datasets. `Generation Type' indicates who generates NL and FOL. Asterisks denote that the output is obtained by a translation (from the NL or the FOL part); no asterisk indicates joint generation.}

\label{tab:dataset-simple}
\end{table}
    
    \paragraph{Generation and validation methods.}
    
    Three main generation paradigms recur across the considered datasets, often in combination: \textit{human generation}, \textit{LLM generation}, and \textit{rule-based generation}. Despite some of the procedures provide a degree of validation \emph{per se} (e.g., human generation), they are sometimes complemented with additional checks.
    
    \emph{Human generation} is used for FOLIO and for the GGC corpus. FOLIO 
    is a NLI dataset consisting of $\sim500$ different stories with a total of $\sim2400$ premises and $\sim1400$ conclusions, written in NL and then formalized to FOL by experts. A first-order theorem prover is used to check whether the entailments are preserved by the FOL formalization, filtering out the instances in which a mismatch occurs.
    GGC collects students' FOL submissions in response to Tarski's World exercises from the textbook \emph{Language, Proof and Logic}~\cite{LPL}, labelled using an automatic grading tool that enforces syntactic and semantic coherence.
    
    \emph{LLM generation} underlies Willow, MALLS, PROOFFOL, and the NL side of ProverQA. Willow 
    extends $\sim300$ human-validated NL–FOL pairs to 16K using GPT‑4, discarding outputs that fail a grammar check; NL–FOL alignment is only partially verified.  
    MALLS 
    collects 28K NL–FOL pairs by prompting GPT-4 to generate NL sentences and then to formalize them using a dynamic $n$-gram counter to encourage diversity. In addition, a human-checked subset of 1K examples is used to identify frequent errors, which are then targeted by rule-based filters to remove such mistakes from the remaining dataset. PROOFFOL 
    uses GPT‑4o to formalize the premises and conclusions of ProofWriter \cite{tafjord-etal-2021-proofwriter}, retaining only instances where the theorem prover's verdict matches the source dataset's ground truth, thereby enforcing both syntactic well-formedness and label consistency.
    ProverQA 
    inverts the direction: an LLM generates NL stories with human-interpretable meanings for predicate symbols, while the FOL side is produced by a symbolic generator, guaranteeing logical structure by construction.
    
    \emph{Rule-based generation} appears in Text2log and in the symbolic part of ProverQA and, in both cases, most validation is intrinsic to the generator.
    Text2log 
    uses predefined grammar rules (through an ad hoc CCG-based semantic parser
    ) to formalize in FOL template-based NL sentences, creating roughly 100K NL–FOL pairs. FOL is correct by construction, but the range of covered NL instances is limited and strongly template-driven.
    
    \paragraph{Discussion.}
    The generation and validation choices have direct implications on how each dataset should be used in autoformalization.

    Datasets with expert-written NL and FOL pairs plus solver-based checks, such as FOLIO and the GGC corpus, provide the strongest available basis for evaluation. Despite the human-conducted generation and the solver-based label-consistency tests, several independent analyses \cite{DBLP:journals/tacl/PeiDJ25,DBLP:conf/acl/YangXPSF24,DBLP:conf/aaai/BrunelloGMMS26} document the presence of annotation errors in FOLIO. In particular, \citet{fixing-folio} found a high rate of ambiguous and incorrect formalizations in the validation split of FOLIO  and in a subset of the MALLS test set; the authors show that these annotation errors can substantially skew reported model performance, and consequently released corrected versions of these datasets for use in evaluating models on this task.

    Large-scale LLM-generated datasets typically guarantee syntactic well-formedness but do not explicitly validate NL–FOL semantic alignment. Systematic error patterns consequently persist, as acknowledged by the authors of MALLS and Willow and further documented in~\citet{DBLP:conf/aaai/BrunelloGMMS26}. A similar consideration applies to rule-based datasets like Text2log, where template-driven NL may not reflect the linguistic variation encountered in realistic settings.

    \subsection{Metrics} \label{sec-metrics}
    Direct evaluation of FOL formalization focuses on comparing a predicted formula $\hat{\varphi}$ with a ground truth formula $\varphi$. We categorize the metrics proposed in the literature into \emph{syntactic} and \emph{semantic}, based on which aspects of the formulas they assess.
    
\subsubsection{Syntactic Metrics}
By \emph{syntactic metrics}, we refer to metrics that treat formulas as syntactic objects, ignoring semantic meaning. A preliminary syntactic validity check ensures that the generated formula conforms to the rules of the formalism; once correctness is verified, several approaches can be applied: \begin{itemize}
    \item \emph{Exact Match} evaluates whether two formulas are identical under string comparison, typically after normalization preprocessing~\cite{journals/corr/abs-2509-22338}.
    \item In \emph{Tree Edit Distance}, instead, formulas are represented as syntax trees, and the minimum-cost sequence of node edit operations to transform one into the other is computed~\cite{DBLP:conf/overlay/BrunelloFGMMS24}.
    \item \emph{BLEU Score}~\cite{papineni2002bleu}, the standard NLP metric, can be applied to FOL formulas after tokenization with an ad-hoc tokenizer tailored for FOL syntax~\cite{DBLP:conf/acl/YangXPSF24}.
    \item \emph{LogicSim} is a metric proposed by  \citet{fernando-lopez-ponce-bel-enguix-2025-limits} as $LogicSim(\varphi, \hat{\varphi}) := pd + tp + ld + IoU$, where $pd$ and $tp$ are the absolute differences in the number of distinct predicates and total predicate occurrences, respectively; $ld$ is the difference in logical operators and quantifiers; and $IoU$ measures the overlap between the predicate sets of $\varphi$ and $\hat{\varphi}$.
\end{itemize}

While easy to compute, syntactic metrics ignore the semantic content of FOL formulas, yielding inadequate results~\cite{DBLP:conf/aaai/BrunelloGMMS26}. Thus, they are often neglected, or used alongside more semantically informed measures, described next.

\subsubsection{Semantic Metrics}
\label{sec-semantic-metrics}

Semantic metrics evaluate whether a candidate formula $\varphi$ and a reference $\hat{\varphi}$ convey the same \emph{meaning}. Direct semantic comparison requires addressing two dimensions in sequence.

\paragraph{Symbol Matching.}
Before any semantic comparison can be made, the symbols of $\varphi$ must be linked to those of $\hat{\varphi}$---an instance of the anchoring problem~\cite{grounding_problem}. Two strategies appear in the literature. The first is \emph{syntactic matching}: \citet{journals/corr/abs-2509-22338} use Levenshtein distance~\cite{Levenshtein1965BinaryCC} to align predicate symbols across signatures by surface-form similarity, a fragile approach that fails when synonymous concepts receive different names or distinct concepts share one. The second strategy is to \emph{provide the signature directly} as part of the task input, as in the LT variants. This eliminates the matching problem by construction and is one of the strongest motivations for the OE/LT distinction of Section~\ref{sec:task-def}.

\paragraph{Formula Closeness.}
Once symbol matching is established, the question is how to measure the semantic distance between the two formulas. In the literature, we can find two metrics.

\emph{Logical equivalence} assigns 1 if and only if $\varphi \equiv \hat{\varphi}$, 0 otherwise: it is theoretically well-founded but binary, thus unsuitable as a possible training objective (see Section~\ref{sec:finetuning}).

\emph{LE score}~\cite{DBLP:conf/acl/YangXPSF24} relaxes this by treating the formulas propositionally~\cite{Smullyan1968-SMUFL} and computing the ratio of truth assignments under which they agree, yielding a continuous value in $[0,1]$.  However, by treating formulas propositionally it discards quantifier structure entirely---precisely what is believed to be the aspect of FOL most challenging to produce correctly.\footnote{More precisely, in the original formulation~\cite{DBLP:conf/acl/YangXPSF24}, symbol matching is not performed independently from formula closeness estimation: the authors select the predicate matching that \emph{maximises} the LE score itself. This conflates the two dimensions and introduces additional shortcomings discussed in~\cite{DBLP:conf/aaai/BrunelloGMMS26}.}

\smallskip

Almost all works implicitly target the \emph{raw translation} variant, treating autoformalization as a single monolithic 
step, an approach that has a direct impact on evaluation. Conflicting empirical results in the literature---some works reporting strong LLM performance on NL--FOL translation~\cite{DBLP:conf/acl/YangXPSF24}, others reporting poor performance~\cite{DBLP:conf/emnlp/HanS0QRZCPQBSWS24}---can be traced to heterogeneous evaluation setups arising from different implicit task formulations \cite{DBLP:conf/aaai/BrunelloGMMS26}. The decomposition introduced in Section~\ref{sec:task-def} addresses these inconsistencies and, together with the metric considerations above, motivates a clear recommendation: when direct semantic comparison is needed, since reliable methods for symbol matching across different signatures are currently lacking, the labelled signature should be provided as input—resolving symbol matching by construction—and \emph{logical equivalence} used as the sole evaluation metric.

\section{LLM-based Autoformalization Methods}
    \label{sec:enhance}
    
    Since current LLMs can 
     be readily prompted to perform FOL-autoformalization, the research has largely focused on techniques for enhancing their performance. A major distinction can be made between methods that fine-tune the LLM (Section~\ref{sec:finetuning}) and those that aim to improve performance without modifying the model’s internal weights, instead relying on prompting strategies (Section~\ref{sec:prompting}) or incorporating a verification stage (Section~\ref{sec:pipelines}).

    Virtually all methods in the FOL-autoformalization literature implicitly target \emph{raw translation} ($p \to \varphi$), and OE-specific methods are largely absent from the literature. As discussed in Section~\ref{sec-metrics}, no metrics are developed for direct evaluation of a labelled signature's quality, making it difficult to optimize or compare OE methods in isolation. This represents a structural gap that we discuss further in Section~\ref{sec:open-gaps}.

\subsection{Fine-tuning Strategies}
\label{sec:finetuning}
Fine-tuning consists of taking a pre-trained model that already encodes broad linguistic and world knowledge and specializing it on a task-specific dataset. Various techniques can be applied, ranging from full-parameter updates to parameter-efficient approaches such as Low-Rank Adaptation, and from purely human-annotated corpora to large-scale synthetic resources.

\paragraph{Supervised Fine-tuning (SFT).}
    The most common strategy updates all model weights, or a subset of them via methods such as Low-Rank Adaptation, using supervised NL–FOL training pairs. Since effective fine-tuning requires a substantial amount of data, most works rely on existing benchmarks. For example, \citet{journals/corr/abs-2509-22338} combine MALLS and Willow to increase diversity and reduce overfitting, while \citet{DBLP:conf/acl/XuWSRYYLQ024} augment data through systematic symbol renaming and NL instructions to promote semantic learning over surface memorization.
    When existing datasets are insufficient, ad hoc datasets can be constructed synthetically. For instance, \citet{DBLP:journals/corr/abs-2206-01962} employ a rule-based pipeline combining established symbolic parsing techniques~\cite{clark-curran-2004-parsing,DBLP:conf/nodalida/Bos15} to convert NL sentences into FOL formulas at scale.

    \paragraph{Reinforcement Learning (RL).}
    Some authors adopt RL in place of, or in addition to, supervised fine-tuning. \citet{DBLP:conf/acl/YangXPSF24} train LogicLLAMA using a combination of the BLEU score and the LE score (Section~\ref{sec-metrics}) as a reward function measuring similarity between the candidate translation $\hat{\varphi}$ and the ground truth $\varphi$, motivated by the observation that a purely autoregressive fine-tuning objective did not yield optimal performance.

    \paragraph{Preference Optimization.}
    Motivated by strong results in other generative settings, preference optimization methods have also been explored.  \citet{fernando-lopez-ponce-bel-enguix-2025-limits} propose a Direct Preference Optimization (DPO) procedure starting from FOLIO, where accepted options correspond to ground-truth formalizations and rejected options to LLM-generated alternatives, augmented with quality scores encoding preference strength.\footnote{The scores derivation procedure is not fully specified.} \citet{DBLP:journals/corr/abs-2506-18383} further study DPO and KTO (Kahneman–Tversky Optimization), finding that KTO after SFT outperforms both SFT alone and SFT with DPO. 

    \subsection{Prompting Strategies}
    \label{sec:prompting}

    Multiple prompting strategies have been explored to enhance the capabilities of pre-trained LLMs in autoformalization without modifying their weights.
    
    \subsubsection{Agnostic Methods}
     In the autoformalization setting, general-purpose prompting techniques such as \emph{Few-Shot prompting}, \emph{In-Context Learning} (ICL), \emph{Chain-of-Thought} (CoT), \emph{Tree-of-Thoughts}, and \emph{Self-Consistency} are commonly used and typically yield performance gains, as shown for instance by~\citet{DBLP:conf/emnlp/HanS0QRZCPQBSWS24}.
   
    \subsubsection{Task-specific Methods}
    Prompting can also be tailored with task-specific instructions (\emph{meta-prompting}): rather than asking the model to produce a formula directly, the prompt decomposes the task into explicit stages.
    For example,~\citet{DBLP:journals/tacl/PeiDJ25,DBLP:conf/acl/HuLX025} first instruct the model to identify a suitable labelled signature (i.e., perform OE) and then carry out the LT step. This staged decomposition is precisely the operationalization of the OE/LT distinction, discussed above. Some works further interpose a theory recovery step:
    \citet{lalwani2024autoformalizing,DBLP:journals/corr/abs-2510-01530,wen-etal-2025-logic} ask the LLM to select among candidate relational axioms before performing translation. \citet{DBLP:conf/iclr/RyuKLY25} decompose the formalization into extracting the logical structure of the sentence, splitting it into simpler components, and performing a sequential translation from subcomponents to the full formula.
    
    Beyond such decompositions, ICL-style methods are also adapted. \citet{DBLP:conf/acl/HuLX025} use retrieval-augmented generation (RAG): a knowledge base of NL–FOL pairs is built from existing datasets and enriched with instances where the model previously failed. Given an input sentence $p$, the closest examples in the knowledge base (according to sentence embeddings) are retrieved and inserted into the prompt as contextual demonstrations.

    \subsection{Output Verification and Refinement Loops}
        \label{sec:pipelines}
        
        Feedback-based methods iteratively refine an LLM's output after an initial formalization attempt. The verification step may rely on formal tools or on the LLM itself, to target either 
        syntactic or semantic correctness.
        
        \subsubsection{Syntactic Verification}
        The most widely adopted form of verification checks whether the generated formula conforms to the target formalization via a syntax checker or theorem prover. Error messages are fed back into the prompt to localize and correct mistakes~\cite{deng2024enhancing,DBLP:conf/emnlp/QuanVDF24}; some works further augment the prompt with a curated list of common errors and fixes~\cite{DBLP:conf/emnlp/PanAWW23,DBLP:journals/corr/abs-2501-14540}. The loop repeats until a valid formula is produced or an iteration limit is reached.

        A complementary approach constrains decoding itself, restricting the output vocabulary at each step to tokens that yield a grammatically valid continuation~\cite{raspanti-etal-2025-grammar}.

        \subsubsection{Semantic Verification}        
        Semantic verification is an open challenge. Assessing whether a FOL formula faithfully captures the meaning of a NL sentence requires bridging two fundamentally different representations, which lies beyond the reach of standard theorem provers. Still, formal tools can  provide partial semantic feedback: when autoformalizing a corpus of NL sentences into a set of axioms~\cite{DBLP:journals/corr/abs-2501-14540}, satisfiability checks can reveal theory-level inconsistencies, and unsatisfiable core extraction can pinpoint offending formulas to guide refinement.

        For richer semantic feedback, several works turn to LLMs themselves. One line of work generates multiple candidate formalizations and aggregates them via self-consistency and majority voting. \citet{DBLP:conf/emnlp/OlaussonGLZSTL23} find that aggregating multiple generations reduces syntactic errors, especially for weaker models, and~\citet{DBLP:conf/aaai/BrunelloGMMS26} observe complementary error patterns across model families, motivating ensembling strategies.

        A second line of work prompts an LLM to directly refine a candidate formalization, like in \citet{fixing-folio}. To mitigate the fallibility of LLM judges, several strategies have been explored: augmenting the context with a knowledge base of NL–FOL pairs including human-corrected formalizations~\cite{DBLP:conf/acl/HuLX025}; prompting the model to compare two candidates and select the better one~\cite{DBLP:journals/corr/abs-2407-02514}; and fine-tuning a dedicated verification model on triplets $(p, \varphi_{\text{pert}}, \varphi)$, where $p$ is the NL sentence, $\varphi_{\text{pert}}$ an incorrect (perturbed) formalization, and $\varphi$ is the ground truth~\cite{DBLP:journals/corr/abs-2409-16461}.
     
        Finally, some approaches combine LLM judgment with formal verification. \citet{DBLP:conf/iclr/RyuKLY25} iteratively compare pairs of candidate formalizations $(\varphi, \varphi')$:  if they are logically equivalent, one is selected arbitrarily;  otherwise an SMT-generated counterexample is passed to the LLM to discriminate between them. \citet{amrollahi_faithful_2026} introduce a roundtrip framework where an SMT solver checks equivalence between a formalization and its back-translation, and an LLM judge diagnoses mismatches to guide targeted repair.

\paragraph{Discussions and takeaways.}
Drawing strong conclusions aimed at answering "Which is the current best approach to FOL-autoformalization?" is difficult for two compounding reasons: the absence of standardized evaluation protocols (and task definition) makes cross-paper comparisons difficult to achieve, and the annotation noise documented in commonly used benchmarks (Section~\ref{sec:datasets}) further limits the significance of performance differences witnessed when using different solutions.

What is reasonable to conclude is that
frontier reasoning models with simple CoT achieve strong results on the LT task \cite{DBLP:conf/aaai/BrunelloGMMS26}; fine-tuning smaller models on specific datasets 
is a viable approach to achieve SOTA performance in a specific context and with less computation (and time) at inference phase. On top of that, prompting techniques (like meta-prompting, decomposing complex sentences into simpler subcomponents, or RAG) can boost the performance of fine-tuned models or already capable ones. 
Finally, adding a verification layer on top of any pipeline seems beneficial as well. 
While syntactic verification represents a robust enhancement across model families, semantic verification addresses a genuinely different and harder problem, making it essential for reliable autoformalization in practice. SAT solvers offer rigorous but partial feedback~\cite{DBLP:journals/corr/abs-2501-14540,amrollahi_faithful_2026}, while LLM-based verification can cover the full semantic dimension but lacks formal guarantees and depends heavily on model capability.

\section{Open Problems and Future Directions} \label{sec:open-gaps}
    We conclude by highlighting open issues in current research and outlining directions for future work.

    \paragraph{On evaluation protocols and benchmarks.} Many evaluation pipelines conflate distinct stages of autoformalization---OE and LT---making it difficult to identify where models struggle. We advocate for standardized evaluation protocols alongside larger, higher-quality benchmarks that are more heterogeneous in domain, formula complexity, and linguistic variation. These benchmarks must also capture overlooked challenges like natural language ambiguity.

    \paragraph{On autoformalization methods.}
    A work on LTL autoformalization has begun to explore structured intermediate representations~\cite{DBLP:journals/access/GerminianiNP25}; analogous strategies for FOL remain preliminary~\cite{202512.1848}. A closer exchange of ideas across formalisms—covering intermediates, post-training strategies, and multi-step pipelines—could substantially advance the field. At the methodological level, almost no existing work explicitly targets OE as a standalone task; developing fine-tuning objectives and prompting strategies that adhere to the OE/LT decomposition, and engaging more directly with the ontology learning literature~\cite{ARMARY2025100693,du2024short}, represents a largely open research agenda. 
    
\paragraph{Towards end-to-end applications.}
The connection between autoformalization research and downstream applications---formal verification, legal reasoning, AI safety---remains largely implicit. End-to-end pipelines that reliably formalize corpora of NL specifications are almost entirely absent. Closing this gap is a crucial next step, allowing concrete application requirements to steer the future direction of the research field.

\section*{Limitations}
This survey does not provide a comparative empirical analysis across enhancement strategies or across LLMs. This is partly by design: as discussed throughout, the absence of standardized evaluation protocols makes cross-paper comparisons unreliable, and one contribution of this work is precisely to clarify the conceptual distinctions---most notably the OE/LT decomposition---that such a standardization would require. By focusing on LLM-based methods, we also set aside earlier symbolic and neural approaches, which are only briefly contextualized in the introduction. Although the surveyed literature evolves rapidly and specific results may become outdated, we believe that the 
contributions of this work 
are largely independent of the specific models and datasets available at any given time, and thus retain their value as the landscape continues to develop.

\bibliography{my_references}

\appendix

\section{First-Order Logic} \label{sec:fol_formalism}
We provide here a brief introduction to First-Order Logic (FOL), covering the concepts needed to follow the main article. Readers seeking a more comprehensive treatment can consult standard textbooks such as \cite{Smullyan1968-SMUFL}.

\subsection{Syntax}
    Two key notions to understand FOL are those of \emph{signature} and \emph{terms}.
\begin{definition}[Signature]
\label{def:signature}
  A signature $\sigma$ is a tuple $(\V,\C,\F,\R)$ such that:
  \begin{itemize}
    \item $\V$ is a (countably infinite) set of \emph{variables};
    \item $\C$ is a set of \emph{constant symbols};
    \item $\F$ is a set of \emph{function symbols}, each with its arity;
    \item $\R$ is a set of \emph{relation/predicate symbols}, each with its arity;
  \end{itemize}
  and $\V \cap \C \cap \F \cap \R = \emptyset$.
\end{definition}

The set of \emph{terms} in FOL is defined inductively as follows:
\begin{enumerate}
  \item every variable in $\V$ and every constant in $\C$ is a term;
  \item if $t_1,\dots,t_n$ are terms and $f \in \F$ is a function symbol
    with arity $n$, then $f(t_1,\dots,t_n)$ is a term.
\end{enumerate}

    \begin{definition}[Syntax of FOL]
    \label{def:fo:syntax}
        An \emph{atomic formula} over the signature $\sigma = \V \cup \C \cup \F \cup \R$ is a string of type $r(t_1,\dots,t_n)$ such that $r \in \R$ is a relation symbol with arity $n$ and $t_i$ is a term, for all $i \in \{1,\dots,n\}$. The set of \emph{FOL formulas $\varphi$ over the signature $\sigma$} is inductively defined as follows:
  \begin{align*}
    \varphi := \ 
         &p(t_1,\dots,t_n)           \ | \  
         \neg \varphi                 \ | \ 
         \varphi \land \varphi            \ | \ 
         \varphi \lor \varphi             \ | \  \\&
         \varphi \to \varphi              \ | \ 
         \varphi \leftrightarrow \varphi  \ | \  
         \exists x \varphi    \ | \
         \forall x  \varphi, 
  \end{align*}
    \end{definition}
where the operators $\neg, \land, \lor, \to, \leftrightarrow$ stand respectively for the negation, the conjunction, the disjunction, the implication, and the equivalence, while $\exists, \forall$ denote the existential and universal quantifiers.

Among the logical symbols mentioned above, parentheses can also be used to clarify which subexpressions a logical connective applies to. For improved readability, parentheses are sometimes omitted; in such cases, the following operator precedence is assumed (from highest to lowest): $\{\lnot,\exists,\forall\}$,  $\land$,
$\lor$, $\to$, $\leftrightarrow$. 

A variable $x$ is said to be \emph{quantified} if it falls within the scope of a quantifier such as $\forall x$ (universal) or $\exists x$ (existential). Otherwise, it is considered a \emph{free variable}. A formula may contain both free and quantified variables: when a formula contains no free variables, it is called a \emph{closed formula} or a \emph{statement}.

Although the syntax described above is the most common, alternative conventions can also be adopted. For instance, one may adhere to the syntactic rules of theorem provers to facilitate interaction with such tools, or encode FOL formulas in a functional style, thereby recasting autoformalization as a code-generation task and exploiting large code-oriented datasets, as in~\cite{DBLP:conf/naacl/Liu25,DBLP:journals/tacl/PeiDJ25}.

In this survey, we do not distinguish among these syntactic variants of FOL, as their differences are largely superficial. It is nonetheless worth noting that~\cite{DBLP:conf/acl-alta/LamTS24} reports that the standard symbolic representation of FOL formulas yields higher performance than alternative encodings.

\subsection{Semantic and SMT solvers}
We now define the semantics of FOL. The core of this definition is the notion of \emph{$\sigma$-structure}, for any signature $\sigma$.

\begin{definition}
  Let $\sigma = \V \cup \C \cup \F \cup \R$ be a signature.
  A $\sigma$-structure $\mathfrak{A}$ is given by:
  \begin{itemize}
    \item a \emph{domain} $\D_{\mathfrak{A}} \neq \emptyset$;
    \item for all $c \in \C$, an element $c_{\mathfrak{A}} \in \D_{\mathfrak{A}}$;
    \item for all $f \in \F$ of arity $n$, a function $f_\mathfrak{A}
      : (\D_\mathfrak{A})^n \to \D_\mathfrak{A}$;
    \item for all $p \in \R$ or arity $n$, a set $p_\mathfrak{A} \subseteq
      (\D_\mathfrak{A})^n$.
  \end{itemize}
\end{definition}

Given a $\sigma$-structure $\mathfrak{A}$ and a variable evaluation $V : \V \to
\D_{\mathfrak{A}}$, for each term $t$, we define $\mathfrak{A}_V(t)$ inductively as follows:
\begin{enumerate}
  \item if $t \in \V$, then $\mathfrak{A}_V(t) = V(t)$;
  \item if $t \in \C$, then $\mathfrak{A}_V(t) = t_\mathfrak{A}$;
  \item if $t = f(t_1,\dots,t_n)$ with $f \in \F$, then $\mathfrak{A}_V(t)
    = f_\mathfrak{A}(\mathfrak{A}_V(t_1),\dots,\mathfrak{A}_V(t_n))$.
\end{enumerate}
In the following, we define the semantics of FOL.

\begin{definition}
  Let $\varphi$ be an FOL formula over the signature $\sigma$, let $\mathfrak{A}$
  be a $\sigma$-structure and let $V$ be a variable evaluation. We define
  the fact that \emph{$\mathfrak{A}_V$ satisfies $\varphi$}, denoted with $\mathfrak{A}_V
  \models \varphi$, inductively as follows:
  \begin{itemize}
    \item $\mathfrak{A}_V \models p(t_1,\dots,t_n)$ iff
      $(V(t_1),\dots,V(t_n)) \in p_\mathfrak{A}$, for all $p \in \R$;
    \item $\mathfrak{A}_V \models \lnot \varphi$ iff $\mathfrak{A}_V \not\models \varphi$;
    \item $\mathfrak{A}_V \models \varphi \land \varphi'$ iff $\mathfrak{A}_V \models \varphi$
      and $\mathfrak{A}_V \models \varphi'$;
    \item $\mathfrak{A}_V \models \varphi \lor \varphi'$ iff $\mathfrak{A}_V \models \varphi$ or
      $\mathfrak{A}_V \models \varphi'$;
    \item $\mathfrak{A}_V \models \varphi \to \varphi'$ iff $\mathfrak{A}_V
      \not\models \varphi$ or $\mathfrak{A}_V \models \varphi'$;
    \item $\mathfrak{A}_V \models \varphi \leftrightarrow \varphi'$ iff $\mathfrak{A}_V
      \models \varphi \to \varphi'$ and $\mathfrak{A}_V \models \varphi' \to \varphi$;
    \item $\mathfrak{A}_V \models \exists x \varphi$ iff there exists a $d
      \in \D_\mathfrak{A}$ such that $\mathfrak{A}_{V'} \models \varphi$, where $V'(x)
      = d$ and $V'(y) = V(y)$ for all $y \neq x$;
    \item $\mathfrak{A}_V \models \forall x  \varphi$ iff for all $d \in
      \D_\mathfrak{A}$ we have that $\mathfrak{A}_{V'} \models \varphi$, where $V'(x) = d$
      and $V'(y) = V(y)$ for all $y \neq x$.
  \end{itemize}
\end{definition}

We say that \emph{$\mathfrak{A}$ satisfies $\varphi$}, denoted with $\mathfrak{A} \models
\varphi$, iff $\mathfrak{A}_V \models \varphi$, for all variable evaluations $V$. Given two FOL formulas $\varphi$ and $\varphi'$ over the signature $\sigma$, we say that \emph{$\varphi$ is equivalent to $\varphi'$} when $\mathfrak{A}_V \models \varphi$ iff $\mathfrak{A}_V \models \varphi'$, for all $\sigma$-structures $\mathfrak{A}$ and all variable assignments $V$: this is the same to say that $\mathfrak{A}_V \models \varphi \leftrightarrow \varphi'$ for all $\sigma$-structures $\mathfrak{A}$ and all variable assignments $V$.

It is worth to notice that these notions, here presented in the general case, become clearer when $\varphi$ is a statement. In this case, since there are no free variables, the fact that $\mathfrak{A}_V \models \varphi$ doesn't depend on the variable assignment $V$. For any two statement $\varphi$ and $\varphi'$, they are \emph{equivalent} when for any $\sigma$-structure $\mathfrak{A}$, we have that $\mathfrak{A} \models \varphi$ iff $\mathfrak{A} \models \varphi'$, or equivalently, $\mathfrak{A} \models \varphi \leftrightarrow \varphi'$.

For every FOL formula $\varphi$, it is possible to transform it into an equivalent formula, over the same signature, such that the negation ($\lnot$) appears only in front of atomic formulas. This normal form is called \emph{Negation Normal Form} (NNF, for short). We define $(\cdot)_{\text{nnf}}$ as the function such that, given in input any FOL formula $\varphi$ it returns its equivalent formula in NNF.

To verify the equivalence between two formulas $\varphi$ and $\varphi'$ over the same signature $\sigma$, we employ SMT (\emph{Satisfiability Modulo Theories}) solvers~\cite{barrett_satisfiability_2018}. These solvers are highly efficient tools for deciding the \emph{satisfiability} of first-order formulas; that is, they determine whether, given a first-order formula $\varphi$ over a signature $\sigma$ as input, there exists a $\sigma$-structure $\mathfrak{A}$ and a variable assignment $V$ such that $\mathfrak{A}_V \models \varphi$.

An SMT solver typically consists of two main components: a Boolean solver that handles the propositional structure of $\varphi$ (often based on the DPLL or CDCL algorithms, as in classical SAT solvers) and a \emph{theory solver} that resolves conjunctions of formulas belonging to a specific theory, such as LRA (Linear Real Arithmetic).
Over the past decades, driven by the remarkable efficiency of SAT solvers, SMT solvers have evolved into powerful tools for automated reasoning, and they now play a crucial role in tasks such as formal verification and planning.

The problem of checking equivalence between two statements $\varphi$ and $\varphi'$ over the signature $\sigma$ is transformed to a satisfiability problem. Specifically, the statements $\varphi$ and $\varphi'$ are equivalent if and only if the statement $\lnot(\varphi \leftrightarrow \varphi')$ is not satisfiable. Formally, $\mathfrak{A} \models \varphi \leftrightarrow \varphi'$ for all $\sigma$-structures $\mathfrak{A}$ if and only if there does not exist a $\sigma$-structure $\mathfrak{A}$ such that $\mathfrak{A} \models \lnot(\varphi \leftrightarrow \varphi')$.

\section{Further NLI Datasets} 
\label{app:reasoning-datasets}

\begin{table*}[t]
\centering
\begin{tabular}{lccc}
\hline
Dataset & NL & FOL & No. of Instances\\
\hline
ReClor \cite{DBLP:conf/iclr/YuJDF20}
& Human
& Not available
& $\sim$6k\\
LogiQA \cite{liu2020logiqa}
& Human
& Not available
& $\sim$8.7k\\
AR-LSAT \cite{DBLP:journals/corr/abs-2104-06598}
& Human
& Not available
& $\sim$2k\\
\hline
\end{tabular}
\caption{Reasoning datasets that require external annotation to be usable for FOL-autoformalization evaluation.}
\label{tab:dataset-ext-annotation}
\end{table*}

\begin{table*}[t]
\centering
\resizebox{\textwidth}{!}{
\begin{tabular}{lccc}
\hline
Dataset & NL & FOL & No. of Instances\\
\hline
RuleTaker/ProofWriter \cite{Clark2020TransformersAS, tafjord-etal-2021-proofwriter}
& Rule-based
& Recoverable
& $\sim$500k\\
PrOntoQA \cite{Saparov2022LanguageMA}
& Rule-based
& Recoverable
& Variable \\
LogicNLI \cite{tian-etal-2021-diagnosing}
& Rule-based
& Recoverable
& $\sim$20k\\
FLD$\times$2 \cite{Morishita2024EnhancingRC}
& Rule-based
& Recoverable
& $\sim$100k\\
\hline
\end{tabular}
}
\caption{Reasoning datasets whose FOL annotation is not shipped as a text field but it is recoverable, with some effort, by inspecting the dataset's generation procedure, in order to be usable for FOL-autoformalization evaluation.}
\label{tab:dataset-rec-annotation}
\end{table*}
As explained in Section \ref{sec:evaluation}, NLI datasets can be used to evaluate FOL-autoformalization (via proxy tasks), provided that the reasoning pattern instantiated in each example can be represented in FOL.
However, some of these datasets were developed for other evaluation purposes and therefore need to be supplemented with additional components before they can be used to evaluate FOL-autoformalization. In the following we will briefly discuss the most recent datasets used in the literature.

Table \ref{tab:dataset-ext-annotation} lists some datasets originally developed to evaluate reasoning in NL, for which no FOL counterpart is available. To make them usable for FOL-autoformalization, an annotator would need to manually formalize the NL sentences that compose each dataset into FOL.
    
Table \ref{tab:dataset-rec-annotation} lists datasets also developed for reasoning tasks, in which the FOL formulas are not shipped as an explicit text field but can be recovered with some effort using rule-based techniques applied to the NL sentences. In several of these datasets (like \citet{tafjord-etal-2021-proofwriter} or \citet{tian-etal-2021-diagnosing}), generation proceeds by first producing a reasoning skeleton in FOL and then translating the FOL formulas into NL via rule-based templates. Since the translation rules are known, this process can in principle be reversed to recover the original FOL formulas even when they are not distributed with the dataset. However, in these cases the mapping between NL sentence and FOL formula is expected to be straightforward, precisely because it is rule-based and template-driven; consequently, it is unlikely to offer a representative picture of the difficulty involved in formalizing a free-form NL sentence.

\end{document}